\documentclass[review,12pt,authoryear]{elsarticle}
\usepackage{graphicx}
\usepackage{xcolor}
\usepackage{amsmath,bm}
\usepackage{multirow}
\usepackage{booktabs}
\usepackage{amsfonts}
\usepackage{tabularx}
\usepackage{algorithm}
\usepackage{algpseudocode}
\usepackage{bm}
\usepackage[hidelinks]{hyperref}
\newcommand{\mname}{FedGDVE}

\usepackage{apalike}
\usepackage{subcaption}  

\newcommand{\rev}[1]{{\color{black}#1}}
\newcolumntype{Y}{>{\centering\arraybackslash}X}

\def\tsc#1{\csdef{#1}{\textsc{\lowercase{#1}}\xspace}}
\tsc{WGM}
\tsc{QE}

\usepackage{apalike}
\journal{Expert Systems with Applications}

\begin{document}

\begin{frontmatter}

\title{Federated Recommender System with Data Valuation for E-commerce Platform}  

\author[label1]{Jongwon Park\fnref{fn1}}
\author[label1]{Minku Kang\fnref{fn1}}
\author[label1]{Wooseok Sim\fnref{fn1}}
\author[label1]{Soyoung Lee\fnref{fn1}}
\author[label1]{Hogun Park\corref{cor1}}\ead{hogunpark@skku.edu}

\fntext[fn1]{These authors contributed equally to this work.}
\cortext[cor1]{Corresponding author.}
\address[label1]{College of Computing and Informatics, Sungkyunkwan University,\\ Suwon, Republic of Korea}

\begin{abstract}
Federated Learning (FL) is gaining prominence in machine learning as privacy concerns grow.
\rev{
This paradigm allows each client (e.g., an individual online store) to train a recommendation model locally while sharing only model updates, without exposing the raw interaction logs to a central server, thereby preserving privacy in a decentralized environment.
Nonetheless, most existing FL-based recommender systems still rely solely on each client's private data, despite the abundance of publicly available datasets that could be leveraged to enrich local training; this potential remains largely underexplored.}
To this end, we consider a realistic scenario wherein a large shopping platform collaborates with multiple small online stores to build a global recommender system. The platform possesses global data, such as shareable user and item lists, while each store holds a portion of interaction data privately (or locally).
\rev{Although integrating global data can help mitigate the limitations of sparse and biased clients' local data, it also introduces additional challenges: simply combining all global interactions can amplify noise and irrelevant patterns, worsening personalization and increasing computational costs.}
\rev{To address these challenges, we propose \textbf{\mname{}}, which selectively augments each client's local graph with semantically aligned samples from the global dataset. \mname{} employs: (i) a pre-trained graph encoder to extract global structural features, (ii) a local valid predictor to assess client-specific relevance, (iii) a reinforcement-learning-based probability estimator to filter and sample only the most pertinent global interactions.}
\mname{} improves performance by up to 34.86\% on recognized benchmarks in FL environments.
\end{abstract}
 
\begin{highlights}
        \item We propose a federated learning method with subgraph-level clients. 

        \item Our proposed model selectively exploits global data to learn local client models.

        \item Our experiments show robust results in both non-IID and IID scenarios. 
\end{highlights}


\end{frontmatter}


\section{INTRODUCTION}
In the contemporary information-abundant environment, the significance of recommender systems cannot be understated, given their ability to curate content tailored to individual tastes ~\citep{wang2023improved, cui2020personalized, wu2021self}.
Collaborative Filtering (CF) has been a fundamental technique in this domain, primarily focusing on analyzing historical user-item interactions to predict personal preferences ~\citep{he2017neural, alharbe2023collaborative}.
A recent development in CF is represented by Graph Collaborative Filtering (GCF), which utilizes the principles of CF with the capabilities of Graph Neural Networks (GNNs) ~\citep{li2023item, wang2022profiling,jung2025balancing,park2025cimage}.
By conceptualizing the interaction dataset as a user-item bipartite graph, GCF harnesses GNNs to craft sophisticated node representations.\
A noteworthy strength of GCF lies in its adeptness at employing an individual user's subgraph structure, enhancing the learning of embeddings by pooling insights from a broad spectrum of users with congruent preferences~\cite{jung2023dual}.
However, this also introduces potential risks related to privacy leakage ~\citep{liu2022neural}.
To address privacy concerns, integrating Federated Learning (FL) within the recommendation domain has emerged as a pivotal development ~\citep{wu2021fedgnn, wahab2022federated}.
FL is a technique to train a machine learning model to benefit from all data dispersed across multiple clients without centralizing, ensuring individual client data remains private.
\rev{Although FL enables privacy-preserving model training, recommender systems in this setting still face a major hurdle: the limited size of each client's local dataset significantly hampers recommendation quality~\citep{DBLP:journals/csur/WangZBYSK24}.}

\begin{figure*}[!t]
    \centering
    \includegraphics[width=0.8\linewidth]{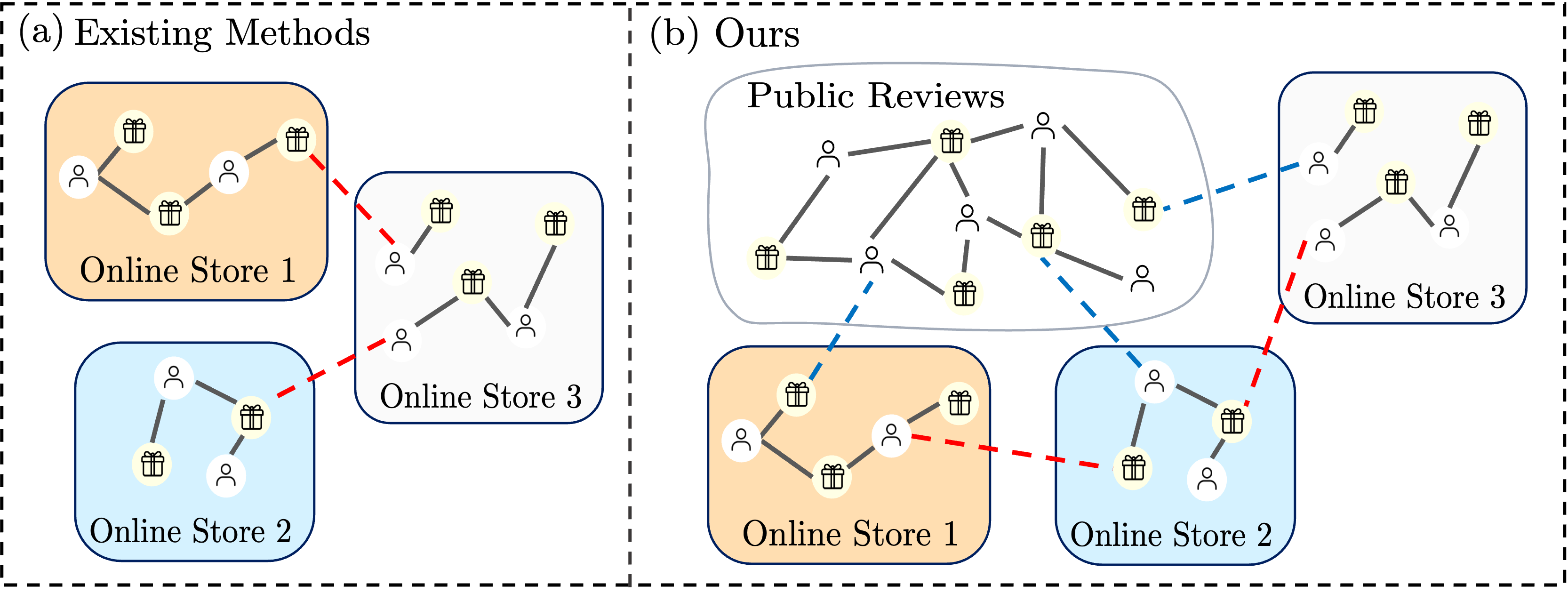}
    \caption{
    \rev{Illustration of a federated recommender system environment with distributed across clients.
    (a) Existing Methods~\citep{peng2022fedni, wu2021fedgnn}: FL-based recommender systems that rely solely on each client's private user-item interaction data without utilizing any external global data. Solid black lines denote observed interactions stored within each client's local dataset, while red dashed lines indicate potential but unrecorded cross-client interactions.
    (b) Ours: \mname{}, identifies and incorporates semantically aligned interactions from a publicly available dataset (e.g., reviews) to support local training. Blue dashed lines represent distributional alignment between local client data and selected global interactions, which are exploited indirectly for training via the graph data value estimator (GDVE) module.}
    }
    \label{fig:useful_scenario}
\end{figure*}

\paragraph{Motivating Scenario}
\rev{Large e-commerce platforms such as Google Shopping~\footnote{https://shopping.google.com}, Shopify~\footnote{https://www.shopify.com}, and Naver Shopping~\footnote{https://shopping.naver.com} operate extensive ecosystems in partnership with many online stores.
Each store seeks to deliver satisfying product recommendations to its customers, yet privacy requirements and country-level regulations~\citep{DBLP:conf/nips/TsoyK23} oblige them to keep customer information strictly private. These constraints make it difficult to train a single recommender on all user-item interactions.
Under these conditions, FL-based recommender systems become a compelling solution: online stores (clients) collaborate by training local models on their private interaction data and share only model parameters with a central server to update a global recommendation model, thereby preserving data privacy.}
\rev{Nevertheless, prevailing FL recommenders~\citep{ammad2019federated, banabilah2022federated, peng2022fedni, yuan2023hetefedrec} are constrained by the local dataset available on each client, which hinders adequate representation learning and yields locally sub-optimal models (Figure~\ref{fig:useful_scenario} (a)).
When such inferior models are aggregated, the gains from collaboration largely evaporate, thereby curbing the overall effectiveness of the federation.}



In real-world e-commerce platforms, however, global data such as product reviews or anonymized logs is often available. These can potentially serve as supplementary training resources if used appropriately.
Our approach aims to take advantage of this possibility. 
\rev{As shown in Figure~\ref{fig:useful_scenario} (b), we incorporate a large, noisy global dataset into each client's training process through selective data utilization.
A naive approach might append the entire global dataset to each client's local data. 
Owing to distributional mismatch and noise, simply merging the entire global dataset into the local data often triggers negative transfer.
The client model is overwhelmed by irrelevant global patterns and drift away from its own distribution.
In addition, processing the entire global dataset imposes significant computational costs, which may be infeasible for resource-constrained clients.
Therefore, there is a clear need for a principled mechanism that can filter and utilize only the most useful parts of the global data.}

To tackle these challenges, we propose the Federated Graph Data Value Estimator (\mname{}), an FL framework for recommender systems that selectively incorporates global interactions aligned with each client's data distribution.
Our foundational setup parallels the work of ~\citep{zhao2018federated}, where global data is available to enhance local models. 
\rev{
Unlike earlier studies that attempted to reinstate IID property by merging each client's data with a small global dataset \citep{zhao2018federated, lv2023fedrds}, \mname{} enhances local training by augmenting each client’s dataset with interactions filtered from a large-scale public dataset that aligns with its own data distribution.
This strategy is therefore far more practical, as it dispenses with the requirement for a carefully curated, limited global dataset and instead operates directly on the large, naturally occurring global knowledge.
}

\rev{
To filter the global dataset effectively, we devise the Graph Data Value Estimator (GDVE), a specialized model designed to identify global interactions that are useful for improving each client’s local model.
GDVE comprises three modules: (i) a graph encoder pre-trained on the global dataset, (ii) a valid predictor pre-trained on the client's local dataset, and (iii) a probability estimator that serves as GDVE's training target.
The graph encoder first transforms each global interaction into latent representations, furnishing rich graph-level features from the global view.
Because these features alone cannot capture client-specific preferences, the valid predictor scores the same interactions to yield a local-belief signal.
The probability estimator then fuses the encoder embeddings with these validity scores to compute a filtering probability for each interaction.
Based on these probabilities, interactions are randomly sampled, enabling client-tailored selection of global data.
However, because the stochastic sampling process is non-differentiable, the probability estimator cannot be optimized directly through backpropagation.
To address this, we introduce an auxiliary task predictor and update the probability estimator using a reinforcement learning scheme.
The task predictor is trained on the subset of global interactions selected by GDVE, and its performance on the client’s local data serves as a reward for the probability estimator.
Interaction subsets that yield higher local performance receive larger rewards, thereby increasing their sampling probabilities in subsequent iterations.
Once GDVE has converged, we proceed to train the recommendation model.
At each training step, the global interactions sampled by GDVE are merged with the client’s local data, allowing the recommendation model to operate on an augmented graph that reflects both private and carefully selected global information.
This selective augmentation unlocks the utility of the vast global dataset while sidestepping its noise and scale.
It steers each local model toward more optimal parameter estimates, and the ensuing refined updates boost the aggregated global model.
Furthermore, by operating on this compact, relevance-filtered subset, \mname{} exploits global data cost-effectively and remains scalable, as it can be applied regardless of the size of the global dataset.
}

 
Our contributions are outlined as follows:
\begin{itemize}
    \item We conduct federated learning in a more realistic yet previously overlooked environment that combines a publicly shared dataset with each client's private local data. To the best of our knowledge, there has been limited exploration in the field of federated recommender systems that considers such a hybrid data setting.
    \item We propose \mname{}, a FL framework that selectively incorporates external data. This approach is designed to enrich user preference information in FL-based recommender systems, particularly in scenarios characterized by sparse local observations.
    \item We observe that \mname{} outperforms recent FL recommender systems in both IID and non-IID settings, with performance gains of up to 34.86\%. 
\end{itemize}

\clearpage
\section{RELATED WORK}
\subsection{Graph neural network for recommender systems}
\rev{Graph recommender systems represent users and items as nodes in a graph, with their interactions as edges to capture an implicit relationship.}
The Graph Collaborative Filtering (GCF) techniques have transitioned from label propagation models such as ItemRank~\citep{gori2007itemrank} to sophisticated graph embeddings, refining the prediction of user preferences. Early models focused on determining structural similarities in user-item interactions but fell short when optimizing against definitive recommendation objectives. Acknowledging the limitations, HOP-Rec ~\citep{yang2018hop} introduced a new blend of graph dynamics and embedding methods, leveraging random walks to broaden user interactions. 
However, this approach also encountered challenges in harnessing higher-order connectivity and was overly dependent on extensive fine-tuning processes.

The advent of models based on representative algorithms, notably STAR-GCN~\citep{zhang2019star} and NGCF~\citep{wang2019neural}, signaled a successful adaptation of GNN to the recommendation domain. 
STAR-GCN uses a reconstruction mechanism to connect multiple adjacent graph blocks with edge masking to avoid privacy problems. 
NGCF, propagating user-item graphs and aggregating information from Graph Convolutional Network (GCN) layers to capture high-order information~\cite{park2020role}, drastically simplified the training process. 
\rev{Our model builds on NGCF by adding a graph-structured contrastive loss, which enhances the learning of implicit relationships in user-item interactions and improves recommendation accuracy in federated settings.}

\subsection{Federated learning for recommender systems}
Federated Learning (FL) is a machine learning technique that promotes collaborative model training across multiple devices while strictly preserving user privacy ~\citep{mcmahan2017communication, banabilah2022federated,kang2025curriculum}.
Contrary to conventional approaches that centralize data, FL prioritizes keeping user data localized while harnessing a centralized vision for improving performance. 
Each device refines its model based on local data and then shares model parameters with a centralized server. 
The server aggregates the shared parameters to enhance a global model, which is redistributed to devices for further refinement ~\citep{chen2019communication, karimireddy2020scaffold}. 

FL has been explored in recommender systems ~\citep{yang2020federated, yi2021efficient, zhao2020privacy}. One noteworthy study is FCF ~\citep{ammad2019federated}, representing the initial attempt to apply FL to standard collaborative filtering methods. FCF demonstrated the stability of recommender system performance within the FL framework. FedMF ~\citep{chai2020secure} extends matrix factorization (MF) by enabling the local learning of user embeddings and transmitting only the gradients to the server. Subsequently, FedMVMF ~\citep{flanagan2021federated} has demonstrated that using multi-view matrix factorization complements the cold-start problem of standard collaborative filters. MetaMF~\citep{lin2020meta} proposed a new MF method to learn the rating prediction in FL using a meta module that reduces the size of the matrix. FedPerGNN ~\citep{wu2022federated} is an FL recommender system using a Graph Neural Network (GNN) and displayed equivalent performance to the centralized model using the neighbor expansion technique with assistance from a third party. 

Recent research has notably emphasized personalized federated learning to address the unique statistical characteristics of each client. For instance, PerFedRec ~\citep{luo2022personalized} clusters clients and maintains the distinct attributes of each group during aggregation for global model parameters. Meanwhile, HeteFedRec ~\citep{yuan2023hetefedrec} underscores the limitations of uniformly distributing global model parameters across clients with heterogeneous resources. As a remedy, it proposes modulating the global parameter size to better align with each client's resources. 

\rev{Our work differs from prior studies in extending the use of GNNs in recommender systems to a subgraph-level FL setting. In this setup, each client corresponds to a group of users and their associated local interaction subgraph. This formulation allows more expressive modeling of user-item relationships within clients, but also intensifies the challenge of limited interaction data available to each client. To address this, we leverage abundant global datasets available online---resources largely overlooked in existing FL-based recommender systems that rely solely on private client data.}






\subsection{Data valuation}
The quantification of data value is significant in the field of machine learning. There is an increasing interest in explaining model predictions and discerning which data to assimilate or exclude to enhance model efficacy \citep{wu2023variance, ghorbani2019data,kang2024unr, park2023generating, jung2025harnessing}. 

One standard method to tackle the challenge of data valuation is utilizing Shapley Value (SV) as the metric to assess the contribution of individual data \citep{mitchell2022sampling, kwon2021beta}. SV defines a profit allocation scheme that complies with a suite of properties, such as fairness, rationality, and decentralizability. The study by \citep{jia2019towards} aimed at reducing computation cost by promoting information sharing amongst diverse model evaluations, given that SV naturally exhibits sparsity, signifying that only a few numbers of data hold substantial values. In a different stride, \citep{ghorbani2020distributional} introduced distributional Shapley, which measures value concerning an overall data distribution instead of a fixed dataset, offering a more dynamic valuation. 
Despite its merits, SV encounters limitations within FL frameworks, where it has difficulty evaluating the value of local data compared to other data sources and fails to consider the sequence in which data is presented.

Another approach to valuing data is integrating reinforcement learning \citep{yoon2020data, tay2022incentivizing}. 
This makes data valuation on large and complex Deep Neural Networks (DNNs) affordable.
A proficient data value estimator can reprioritize the use of training samples and enable the training of high-performance predictors with noisy or out-of-domain training samples. The training is based on the predictor's performance on a small validation data set. A batch of training samples is given as input, yielding a selection probability that effectively ranks the batches by importance.

\rev{Our work builds on these ideas by applying a reinforcement learning–based data valuation strategy specifically to filter large, noisy global datasets before augmenting each client’s local data in an FL-based recommender system. This selective filtering ensures that only semantically relevant global interactions are incorporated, mitigating noise and reducing computational overhead.}

\section{METHOD}

\subsection{Notations}

Throughout this section, we define the sets of users and items using the notations $\mathcal{U} = \{u_1, u_2, \ldots, u_P\} $ and $\mathcal{T} = \{t_1, t_2, \ldots, t_Q\}$, respectively. 
Here, $P$ and $Q$ denote the total number of users and items. We consider the interactions between users and items, such as reviews or ratings, to construct a bipartite graph $G$, which is represented by an adjacency matrix $\mathbf{A} \in \{0,1\}^{P \times Q}$. 
In this matrix, 1 indicates the presence of an interaction between a user and an item, while 0 signifies no interaction. 
Furthermore, each user and item is associated with an embedding vector of dimension \(d\).
\rev{These embedding vectors are denoted by \(\mathbf{e_{\text{user}}}(i) \in \mathbb{R}^{d}\) for the \(i\)-th user and \(\mathbf{e_{\text{item}}}(j) \in \mathbb{R}^{d}\) for the \(j\)-th item, where \(\mathbf{E_{\text{user}}} \in \mathbb{R}^{P \times d}\) and \(\mathbf{E}_\text{item} \in \mathbb{R}^{Q \times d}\) are matrices containing the embedding vectors for all users and items, respectively.}

Consider $K$ clients participate in the training.
Client $k$ holds an internal local data set $D_{\text{local}}^{k}$ that is not shared with other clients.
This dataset comprises interactions between $P^k_{\text{local}}$ users and $Q^k_\text{local}$ items specific to client $k$.
The collection of all clients' private datasets is denoted by  \( D_{\text{local}} = \bigcup_{k=1}^{K} D_{\text{local}}^{k}  \), where \( \bigcup \) signifies the union of all private local datasets. 
In contrast, the platform maintains a global dataset \( D_{\text{global}} \), which is available to all clients. 
\rev{The global dataset contains $P_{\text{global}}$ users and $Q_{\text{global}}$ items.
We assume it is larger in scale than any individual local dataset, such that $P_{\text{global}} \geq P^k_{\text{local}}$ and $Q_{\text{global}} \geq Q^k_{\text{local}}$ for every client $k \in {1,\ldots,K}$
The distributions of $D_{\text{local}}$ and $D_{\text{global}}$ are not necessarily identical.}

\begin{figure}[t]
    \centering
    \includegraphics[width=1\linewidth]{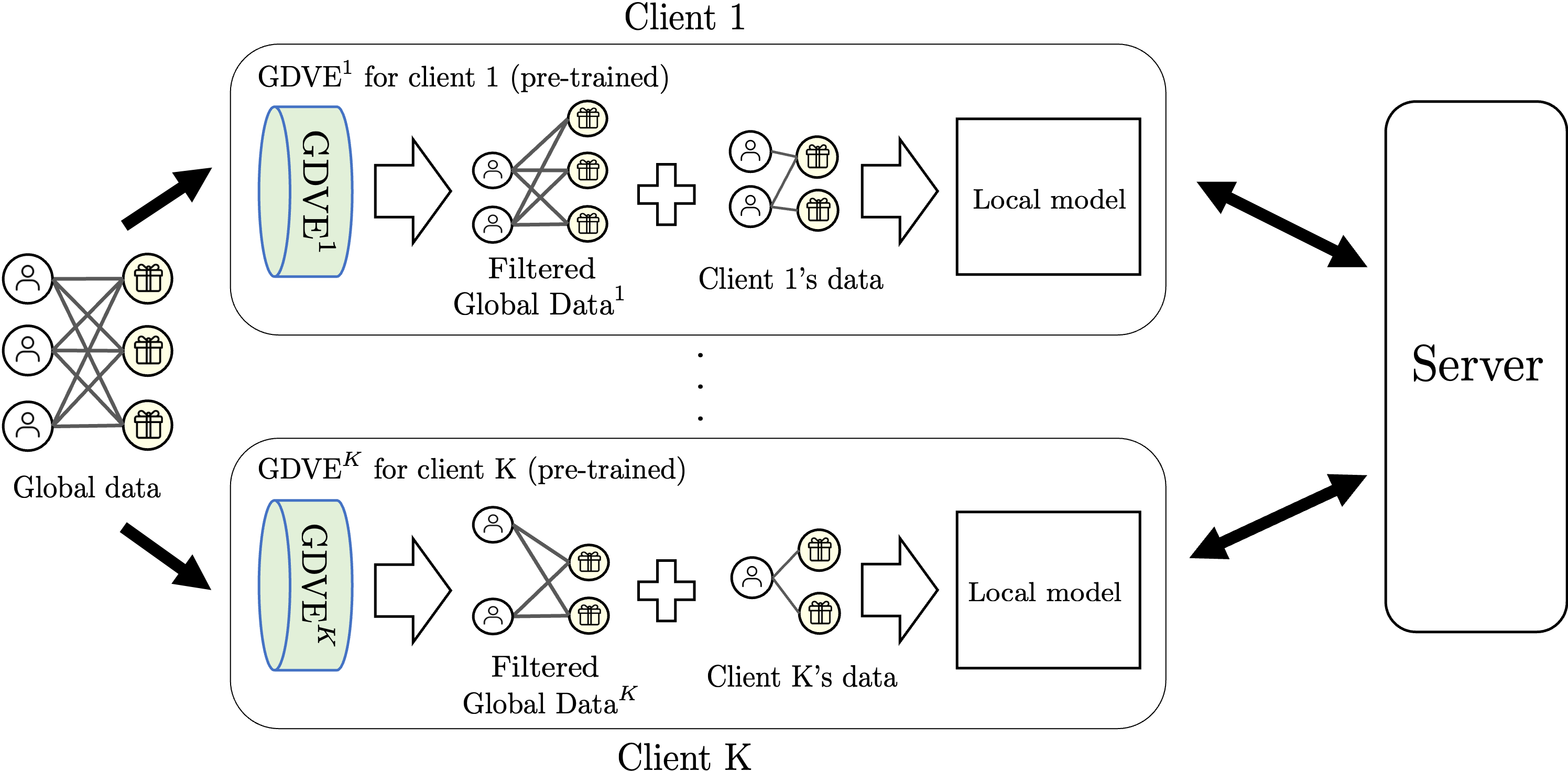}
    \caption{\rev{Overview of \mname{}.}}
    \label{fig:Fed-GDVE}
\end{figure}

\vspace{12mm}
\rev{\subsection{Overview of the \mname{} framework}}

\begin{figure*}[t]
    \centering
    \includegraphics[width=1.0\linewidth]{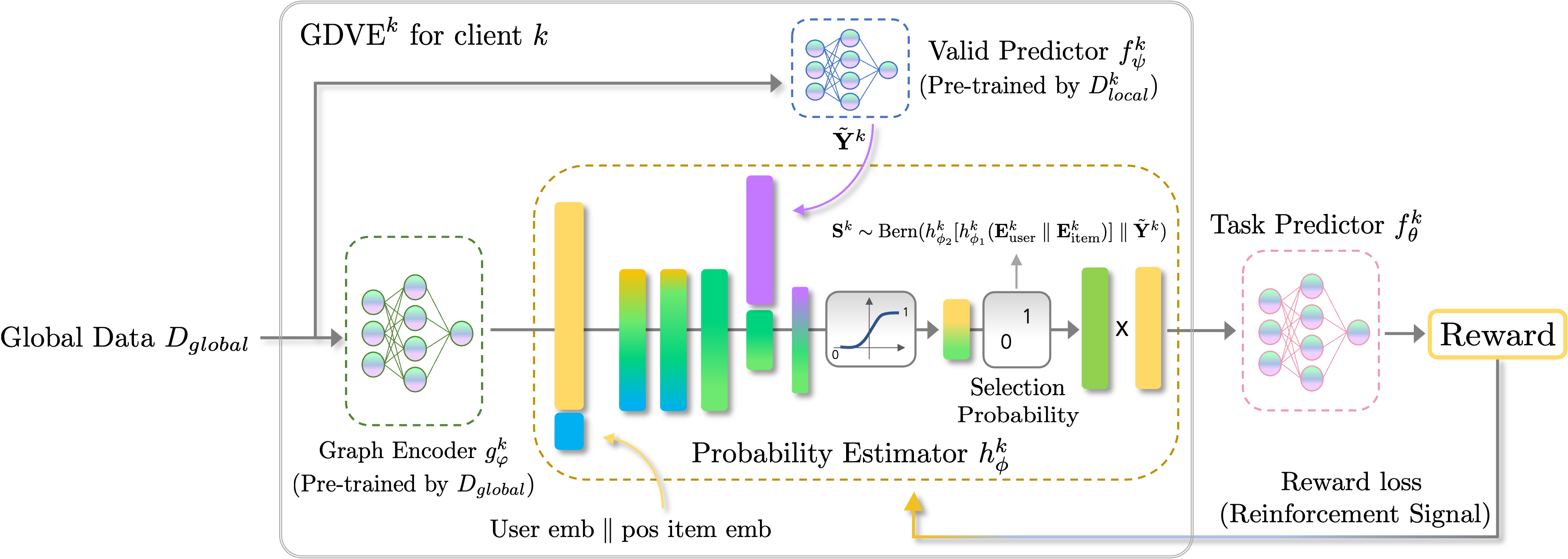}
    \caption{\rev{The overall architecture of the proposed $\text{GDVE}^{k}$ for client $k$.}}
    \label{fig:GDVE}
\end{figure*}

\rev{An overview of the \mname{} framework is presented in Figure~\ref{fig:Fed-GDVE}.
As shown in the figure, client $k$ maintains both a GDVE model and a local recommendation model.
The GDVE is retained locally within each client and is never transmitted to the server.
It filters interactions from $D_\text{global}$ that are useful for the client's data.
The filtered interactions are used to augment the local dataset $D_{local}^k$, enabling each client to utilize global data more effectively than naively incorporating it in its entirety.
This augmented dataset is then used to train the local recommendation model, which can be any arbitrary recommendation architecture.
Once training is complete, the model parameters are transmitted to the server for aggregation.}

\rev{\subsection{GDVE Architecture}}
\rev{In this section, we describe the overall architecture of GDVE.
As illustrated in Figure~\ref{fig:GDVE}, the GDVE for client $k$ comprises three modules: a graph encoder $g_{\varphi}^{k}$, a valid predictor $f_{\psi}^{k}$, and a probability estimator $h_{\phi}^{k}$.
The graph encoder and the valid predictor share the identical NGCF backbone structure \citep{wang2019neural}.

Specifically, the graph encoder for client $k$ is pre-trained on the global interaction graph $D_{global}$, mapping global user-item interactions into a latent space and yielding the embedding matrices $\mathbf{E}^k_{user}$ and $\mathbf{E}^k_{item}$.
Reinforcement learning–based data value estimators typically rely on explicit feature vectors to assign a selection probability to each data point \citep{yoon2020data, tay2022incentivizing}.
In recommender systems, however, individual user–item interactions lack such features, rendering a direct application of these estimators infeasible.
The embeddings generated by the graph encoder serve as surrogate feature vectors that encode global interaction patterns, enabling reinforcement learning-based data value estimation to function effectively within a GCF-based recommender system.

Nevertheless, the representations produced by the graph encoder alone remain insufficient to fully capture the client's specific preferences toward the global dataset.
To this end, the valid predictor---pre-trained on the client's local dataset---assigns a validity score for each global interaction.
These scores are collectively represented as the matrix $\tilde{\mathbf{Y}}^k$, where a higher value indicates a stronger client preference for the corresponding interaction.

Finally, the probability estimator combines the outputs of the graph encoder and the valid predictor to compute the selection probability for each interaction, after which interactions are stochastically sampled according to these probabilities.
It is implemented as two sequential Multi Layer Perceptron (MLP), $h^k_{\phi_1}$ and $h^k_{\phi_2}$.
The first network, $h^k_{\phi_1}$, transforms the embeddings into an intermediate representation, while the second, $h^k_{\phi_2}$, concatenates this representation with the associated validity score and maps the result to a scalar selection probability $\mathbf{P}^k$ for each interaction.
Based on these probabilities, client $k$ derives a binary mask $\mathbf{S}^k$ to filter the global interactions.}

\rev{\subsection{Learning Procedure of GDVE}}

\rev{In this section, we detail the learning procedure of GDVE.
The complete learning pipeline is presented in Algorithm~\ref{alg:GDVE}.}

\rev{\subsubsection{Pre-training of graph encoder and valid predictor}}

\rev{Prior to training the probability estimator, client $k$ pre-trains the graph encoder $g_{\varphi}^{k}$ on the global dataset $D_{global}$ and the valid predictor $f_{\psi}^k$ on the client's local dataset $D_{local}^k$.
Once this pre-training phase is complete, the parameters of both modules are kept frozen.
Both modules are optimized using the Bayesian Personalized Ranking (BPR) loss \citep{DBLP:journals/corr/abs-1205-2618}.
The BPR loss is a specialized ranking loss function tailored for recommender systems, especially when dealing with implicit feedback. 
It derives implicit negative feedback from unobserved interactions and optimizes the model to rank observed items higher than unobserved ones. 
This pairwise strategy effectively captures user preferences and has been shown to deliver superior performance in settings where only positive interactions are available.
Formally, the BPR loss is defined as follows:}

\rev{\begin{equation}
\mathcal{L}_{\text{BPR}} = -\sum_{u \in U} \sum_{m \in T} \sum_{n \notin T} \ln \sigma(\hat{y}_{um} - \hat{y}_{un}) + \lambda \| \mathbf{W} \|_2^{2},
\label{eq:bpr_loss}
\end{equation}
where $U$ denotes the set of users and $T$ denotes the set of items.
Concretely, the BPR loss for the graph encoder is instantiated with the user and item sets of $D_{\text{global}}$.
In contrast, the BPR loss for the valid predictor uses the corresponding user and item sets of the client $k$'s local dataset $D^k_{\text{local}}$.
$\hat{y}_{um}$ indicates the preference score for user $u$ and item $m$, computed as the matrix multiplication of their corresponding embeddings predicted by the model.
$\sigma$ is the sigmoid function, $\lambda$ is the regularization coefficient, and $\mathbf{W}$ represents the model parameters.
The first term, $\ln \sigma(\hat{y}_{um} - \hat{y}_{un})$, encourages the model to assign higher scores to observed interactions than to unobserved ones.
This mechanism enables a nuanced modeling of user preferences and item affinities, effectively tailoring the learned embeddings to reflect the underlying interaction patterns.
The second term $\lambda \| \mathbf{W} \|_2^{2}$ introduces regularization to mitigate overfitting by penalizing the magnitude of the parameters.}

\rev{\subsubsection{Task predictor}}


\rev{As depicted in Figure~\ref{fig:GDVE}, the probability estimator $g_{\varphi}^{k}$ of client $k$ is updated by a reward guided by an auxiliary task predictor.
To clarify the role of the task predictor, we first outline how GDVE derives the interaction mask $\mathbf{S}^k$ for client $k$.
GDVE processes each mini-batch $\mathcal{B}$ consisting of $\beta$ users sampled from the global dataset, rather than utilizing the entire $D_{\text{global}}$.
For each mini-batch, the graph encoder $g_{\varphi}^{k}$ produces user and item embedding matrices $\mathbf{E}^k_{\text{user}} \in \mathbb{R}^{\beta \times d}$ and $\mathbf{E}^k_{\text{item}} \in \mathbb{R}^{Q_{\text{global}} \times d}$.
Similar to the graph encoder, the valid predictor $f_\psi^k$ outputs user and item embedding matrices $\tilde{\mathbf{E}}^k_{\text{user}} \in \mathbb{R}^{\beta \times d}$ and $\tilde{\mathbf{E}}^k_{\text{item}} \in \mathbb{R}^{Q_{\text{global}} \times d}$, respectively.
The validity score matrix $\tilde{Y}^k \in \mathbb{R}^{\beta \times Q_{\text{global}}}$ is subsequently computed as the matrix multiplication of the user and item embeddings.
The probability estimator derives the interaction mask $\mathbf{S}^k \in \{0,1\}^{\beta \times Q_{\text{global}}}$ as follows:

\begin{align}
\mathbf{P}^k &= h^k_{\phi_2} \left( \left[ h^k_{\phi_1} \left( \mathbf{E}_{\text{user}}^k \, \| \, \mathbf{E}_{\text{item}}^k \right) \right] \, \| \, \tilde{\mathbf{Y}}^k \right), \\
\mathbf{S}^k &= \mathrm{Bern}\left( \mathbf{P}^k \right),
\label{eq:mask}
\end{align}
where $\mathbf{P}^k \in \mathbb{R}^{\beta \times Q_\text{global}}$ indicates selection probability , $\|$ denotes concatenation, and $\mathrm{Bern}$ represents the Bernoulli distribution.
The probability estimator stochastically selects the interactions used for training recommendation model according to their estimated probabilities.

However, the sampling operation of the probability estimator is non-differentiable, precluding direct optimization via gradient descent.
We therefore update it indirectly with a reward signal derived from downstream performance on validation data.
Serving as an intermediary, the task predictor $f_\theta^k$, which adopts the NGCF backbone, estimates this reward and forwards it to the probability estimator.
The task predictor is first optimized on the subset of interactions selected by $\mathbf{S}^k$, minimizing the following objective:

\begin{equation}
\mathcal{L}_{\theta}^k = \mathcal{L}_{\text{BPR},\theta}^k + \mathcal{L}_{\text{struc}}^k,
\label{eq:tp_loss}
\end{equation}
where $\mathcal{L}_{BPR,\theta}^k$ refers to the BPR loss with respect to $\theta$, as defined in Equation~\eqref{eq:bpr_loss}.
In addition, to make the reward more discriminative, we incorporate the structure loss $\mathcal{L}_{\text{struc}}^k$, which is formulated as a contrastive objective following the InfoNCE \citep{oord2018representation} to encourage structurally informed representations.
Specifically, let $\mathbf{e}_\text{user}^{(0)}(i)$ be the initial embedding of the $i$-th user and $\mathbf{e}_\text{user}^{(l)}(i)$ the embedding obtained from an even-numbered GNN layer $\ell \in \mathcal{E}$.
For every user $i$ in the current mini-batch $\mathcal{B}$ we treat $\left(\mathbf{e}_\text{user}^{(0)}(i),\mathbf{e}_\text{user}^{(l)}(i) \right)$ as a positive pair, while $\left(\mathbf{e}_\text{user}^{(0)}(i),\mathbf{e}_\text{user}^{(l)}(j) \right)$ with $j \neq i$ forms a negative pair.
The resulting structure loss is

\begin{equation}
    \label{eq:struct_loss}
    \mathcal{L}_{\text{struc}}^{k}
= -\sum_{i \in \mathcal{B}}
    \log
    \frac{\displaystyle\sum_{\ell \in \mathcal{E}}
          \exp \bigl(\langle \mathbf{e}_{\text{user}}^{(0)}(i),\,
                           \mathbf{e}_{\text{user}}^{(\ell)}(i) \rangle / \tau\bigr)}
         {\displaystyle\sum_{\ell \in \mathcal{E}}\!
          \sum_{j \in \mathcal{B}}
          \exp \bigl(\langle \mathbf{e}_{\text{user}}^{(0)}(i),\,
                           \mathbf{e}_{\text{user}}^{(\ell)}(j) \rangle / \tau\bigr)},
\end{equation}
where $\tau$ is a temperature and $\langle \cdot,\cdot\rangle$ denotes the inner product.
This contrastive loss pulls together a user's embeddings across layers while pusing apart those of different users within the batch $\mathcal{B}$.
Without structure loss, the task predictor trained solely with the BPR loss is prone to overfitting sparse and biased interactions, which inflates reward noise and destabilizes the sampling policy.
Structure loss regularizes the predictor by aligning each user's embeddings across GNN layers while pushing apart those of unrelated users, thereby injecting higher-order structural cues that the BPR loss alone cannot capture.
This sharpening effect renders the reward more discriminative, so interactions that align with the client’s underlying graph structure consistently receive larger rewards, whereas structurally irrelevant or noisy interactions are down-weighted.}


\rev{\subsubsection{Training of probability estimator}}

\rev{After training the task predictor, the performance of the client $k$'s predictive model $f_\theta^k$ on the local data $D_{\text{local}}^k$ is employed as a reward $\mathcal{R}_h^k$ to update the parameters of the probability estimator $\phi$ through a policy-gradient step $\nabla_\phi l(\phi)$, as detailed in the following equation.

\begin{align}
\label{eq:policy}
&
\nabla_\phi l(\phi) = 
\mathbb{E}\Bigg[
\mathcal{R}_h^{k}\!\bigl(f_{\theta}^{k}(D_{\text{local}}^{k})\bigr)\; \times \nonumber \\
& \quad\quad\quad\quad\quad
\frac{1}{|\mathcal{B}|}
\sum_{b\in\mathcal{B}}
\nabla_\phi\;
\Bigl\langle
\mathbf S^k_{b}\odot\log\mathbf P^k_{b}
+\bigl(1-\mathbf S^k_{b}\bigr)\odot\log\!\bigl(1-\mathbf P^k_{b}\bigr),
\;\mathbf 1
\Bigr\rangle
\Bigg],
\end{align}
where $\mathcal{B}$ denotes the mini-batch of users, $\mathbf{P}_b^k$ is the selection probability vector for user $b$, $\mathbf{S}_b^k$ is the corresponding binary interaction mask, and $\mathbf 1$ is a vector of ones used to aggregate the element-wise log-likelihoods across items.
Intuitively, the reward $\mathcal{R}_h^k$ modulates the magnitude---and hence the direction---of the policy-gradient update.
When the task predictor achieves stronger validation performance, $\mathcal{R}_h^k$ becomes larger, amplifying the gradient term and increasing the likelihood that the probability estimator will reinforce the current sampling pattern of global interactions.
Conversely, a weaker reward down-weights the update, discouraging policies that fail to improve the client's local objective.
In this way, GDVE progressively concentrates its sampling probability on interaction combinations that demonstrably boost downstream performance on local data.
After the training of GDVE is completed, it remains unsynchronized with other GDVEs to preserve the unique properties of each local dataset.}

\begin{algorithm}
\caption{\textsc{GDVE} training on client $k$}
\label{alg:GDVE}
\begin{algorithmic}[1]
\State \textbf{Inputs:} global data $D_{\text{global}}$ (user set $\mathcal U_{\text{global}}$, item set $\mathcal T_{\text{global}}$), local data $D_{\text{local}}^{k}$; learning rates $\eta$ and $\gamma$; mini-batch size $\beta$
\State \textbf{Initialize:} task-predictor parameters $\theta$, probability estimator parameters $\phi$

\State Pre-train graph encoder $g_{\varphi}^{k}$ on $D_{\text{global}}$ \hfill$\triangleright$ global structure
\State Pre-train valid predictor $f_{\psi}^{k}$ on $D_{\text{local}}^{k}$ \hfill$\triangleright$ local preference
\While{not converged}
    \State Sample mini-batch $\mathcal B\subset\mathcal U_{\text{global}}$ of $\beta$ users
    \State $\mathcal L_{\text{batch}}\leftarrow 0$
    \For{\textbf{each} user $u\in\mathcal B$}
        \State $(\mathbf e_{\text{user}}^{k}(u),\mathbf E_{\text{item}}^{k}) \gets g_{\varphi}^{k}(u,\mathcal T_{\text{global}})$ \hfill$\triangleright$ user and item embeddings
        \State $(\tilde{\mathbf e}_{\text{user}}^{k}(u),\tilde{\mathbf E}_{\text{item}}^{k}) \gets f_{\psi}^{k}(u,\mathcal T_{\text{global}})$
        \State $\tilde{\mathbf Y}^k
               = \tilde{\mathbf e}_{\text{user}}^{k}(u)\,\tilde{\mathbf E}_{\text{item}}^{k\top}$ 
               \hfill$\triangleright$ validity score
        \rev{
        \State $\mathbf P_{u}^{k}
               = h_{\phi_2}^{k}\bigl([\,
                 h_{\phi_1}^{k}(\mathbf e_{\text{user}}^{k}(u)\parallel\mathbf E_{\text{item}}^{k})
                 ]\parallel\tilde{\mathbf Y}^k\bigr)$
        \State $\mathbf S_{u}^{k}\sim\mathrm{Bern}(\mathbf P_{u}^{k})$ \hfill$\triangleright$ global interaction mask
        \State $\mathcal L_{\text{batch}} \leftarrow \mathcal L_{\text{batch}}
               + \bigl(\mathcal L_{\text{BPR},\theta}^{k}(u)
               + \mathcal L_{\text{struc}}^{k}(u)\bigr)$
        }
    \EndFor
    \State Update $\theta$ with batch-average loss:
    \rev{
    \State \hspace{1.5em}$\displaystyle
    \theta \leftarrow \theta - \frac{\eta}{\beta} \nabla_\theta \bigl( \mathcal{L}_{\text{batch}} \bigr)
    $
    }
    \rev{\State Compute reward $r^{k} = \mathcal R_{h}^{k}\bigl(f_{\theta}^{k}(D_{\text{local}}^{k})\bigr)$}
    \State Policy update for $\phi$:
    \State \hspace{1.5em}$\displaystyle
           \phi \leftarrow \phi-\gamma\,r^{k}\,
           \nabla_{\phi}l(\phi)$
\EndWhile
\end{algorithmic}
\end{algorithm}

\rev{\subsection{Training recommendation model}}

\rev{
Client $k$'s GDVE first filters the global interactions that are expected to be most beneficial for improving the performance of its local model trained on $D_\text{local}^k$.
To keep the large-scale global data from overwhelming local information, users in global dataset are sampled in mini-batches.
The interactions retained by GDVE are then used to augment the local dataset, and the local recommendation model is trained on this augmented dataset using both the BPR loss (Equation~\eqref{eq:bpr_loss}) and the structure loss defined in Equation~\eqref{eq:struct_loss}.
}

After the local training phase, the parameters of the local models are transmitted to the server and aggregated through a weighted average based on the original size of each local dataset~\citep{mcmahan2017communication}.
Subsequently, the updated parameters of the global model are dispatched back to the clients.
Each client integrates these shared parameters and resumes its local training.
This iterative cycle repeats until the global models stabilize and converge.

\section{EXPERIMENT}
\subsection{Experimental Setting}
\begin{table}[!t]
\centering
\caption{Summary of datasets.}
\resizebox{0.7\linewidth}{!}{%
\begin{tabular}{lcccc}
\toprule
\textbf{Dataset} & \textbf{\# of users} & \textbf{\# of items} & \textbf{\# of edges} & \textbf{Density} \\
\midrule
ml-100k & 943 & 1,682 & 99,975 & \rev{6.303\%} \\
Gowalla & 29,858 & 40,981 & 1,027,370 & \rev{0.084\%} \\
Yelp2018 & 31,668 & 38,048 & 1,561,406 & \rev{0.013\%} \\
\bottomrule
\end{tabular}
}
\label{tab: summary}
\end{table}

In this work, we conducted experiments using \text{MovieLens-100k} \citep{herlocker1999algorithmic}, \text{Gowalla} \citep{liu2013personalized}, and \text{Yelp2018} \citep{wang2019neural} dataset benchmarks.
Table \ref{tab: summary} shows the summary of the datasets used. For data split, only NGCF \citep{wang2019neural} uses a 100\%  centralized split. All other FL models use 50\% as global data, and the remaining 50\% is equally distributed between 10 clients.

Two models were selected as baselines for comparison: \text{FedMF} \citep{chai2020secure} and \text{FedPerGNN} \citep{wu2022federated}. The \text{FedMF} serves as a fundamental approach to federated recommender systems. In contrast, \text{FedPerGNN}, a more contemporary model that has demonstrated commendable performance, stands as one of the most recent advancements in this field. 
For hyperparameters, we set temperature $\tau$ as 0.1, batch size $B$ as 1024, learning rates $\eta$ as 0.004 and $\gamma$ as 0.007. In the architecture of our proposed model, the graph encoder, valid predictor, and task predictor are each configured with three graph convolution layers. These layers are uniformly initialized by an embedding dimensionality of 64. The probability estimator consists of three fully connected layers with an embedding dimension of 50, followed by two layers with sizes [30, 1].

\vspace{3mm}
\subsection{\rev{Experimental Results and Analysis}}
\paragraph{\rev{Overall Performance Comparison}}
\begin{table}[]
\centering
\caption{Comparative performance analysis on three benchmark datasets. \textbf{P}, \textbf{R}, and \textbf{N} denote precision, recall, and NDCG respectively. In each column, the \textbf{boldfaced} score denotes the best result among the FL baselines, and the \underline{underlined} score represents the second-best result among the FL baselines. We note that our model introduces a federated learning (FL) method, building upon the foundation of NGCF.}
\resizebox{\textwidth}{!}{%
\begin{tabular}{@{}lccccccccc@{}}
\toprule
  & \multicolumn{3}{c}{\textbf{MovieLens-100k}} & \multicolumn{3}{c}{\textbf{Gowalla}} & \multicolumn{3}{c}{\textbf{Yelp2018}} \\
\cmidrule(lr){2-4}
\cmidrule(lr){5-7}
\cmidrule(lr){8-10}
\textbf{Methods} & \textbf{P@100} & \textbf{R@100} & \textbf{N@100} & \textbf{P@100} & \textbf{R@100} & \textbf{N@100} & \textbf{P@100} & \textbf{R@100} & \textbf{N@100} \\ \midrule
NGCF (Centralized) & 0.0864 & 0.5368 & 0.3252 & 0.0187 & 0.2990 & 0.1488 & 0.0152 & 0.1654 & 0.0816 \\ \midrule
FedMF & 0.0581 & 0.3784 & 0.1787 & 0.0152 & 0.2336 & 0.1235 & 0.0067 & \textbf{0.1323} & 0.0498 \\
FedPerGNN & 0.0641 & 0.4609 & 0.2317 & 0.0201 & 0.3257 & 0.1672 & 0.0088 & 0.0874 & 0.0425 \\
FedNGCF & 0.0630 & 0.4573 & 0.2266 & 0.0194 & 0.3120 & 0.1583 & 0.0113 & 0.1184 & 0.0585 \\ \midrule
\textbf{FedGDVE (-GDVE)} & 0.0626 & 0.4596 & 0.2248 & 0.0196 & 0.3172 & 0.1593 & 0.0096 & 0.1027 & 0.0501 \\
\textbf{FedGDVE (-SL)} & \underline{0.0805} & \underline{0.5040} & \underline{0.2930} & \textbf{0.0221} & \underline{0.3296} & \underline{0.1800} & \underline{0.0114} & 0.1205 & \underline{0.0592} \\ \midrule
\textbf{FedGDVE} & \textbf{0.0850} & \textbf{0.5225} & \textbf{0.3115} & \underline{0.0213} & \textbf{0.3363} & \textbf{0.1819} & \textbf{0.0139} & \underline{0.1253} & \textbf{0.0631} \\ \bottomrule
\end{tabular}%
}
\label{tab:comparison}
\end{table}

Table \ref{tab:comparison} presents a comparative analysis of \mname{}'s performance with other established recommendation models. This study employs three widely recognized performance measures for top-k recommendation: precision, recall, and NDCG. Our analysis is centered on the performance associated with the top 100 recommended items. When considering the global data without any selection, FedPerGNN and FedNGCF exhibit commendable performance across all indicators. This suggests that considering neighbor embeddings enhances performance compared to solely relying on client-specific data. 
In contrast, \mname{}, which integrates GDVE with a structural loss component to selectively leverage global data, significantly outperforms baseline models. Specifically, for the MovieLens-100k dataset, \mname{} achieves a substantial increase of 34.86\% in precision, 14.25\% in recall, and 34\% in NDCG compared to FedNGCF, 
\rev{while achieving over 90\% of the centralized NGCF's performance, validating its practicality under privacy constraints.} Enhancements are evident with the Gowalla dataset, where \mname{} achieves improvements of 9.89\% in precision, 7.77\% in recall, and 14.85\% in NDCG. Additionally, the efficacy of \mname{} is demonstrated on the larger Yelp2018 dataset, which is significantly more extensive, as indicated by a 51.98\% increase in density compared to Gowalla. On this dataset, \mname{} achieves a 23.01\% enhancement in precision, a 5.83\% increase in recall, and a 7.86\% improvement in NDCG metrics. \rev{These consistent improvements across datasets of varying sizes and sparsity levels underscore the scalability and robustness of \mname{} in practical federated environments.}
However, it is observed that \mname{}(-GDVE), which excludes the GDVE component and relies solely on structural loss, exhibits only a marginal impact on performance. \rev{This result highlights that contrastive structural learning alone is insufficient for effectively aligning heterogeneous local and global distributions and reinforces the critical role of GDVE in facilitating meaningful global knowledge transfer.}

\paragraph{\rev{Convergence and Performance Comparison}}
\begin{figure}[!t]
    \centering
    \includegraphics[width=0.7\linewidth]{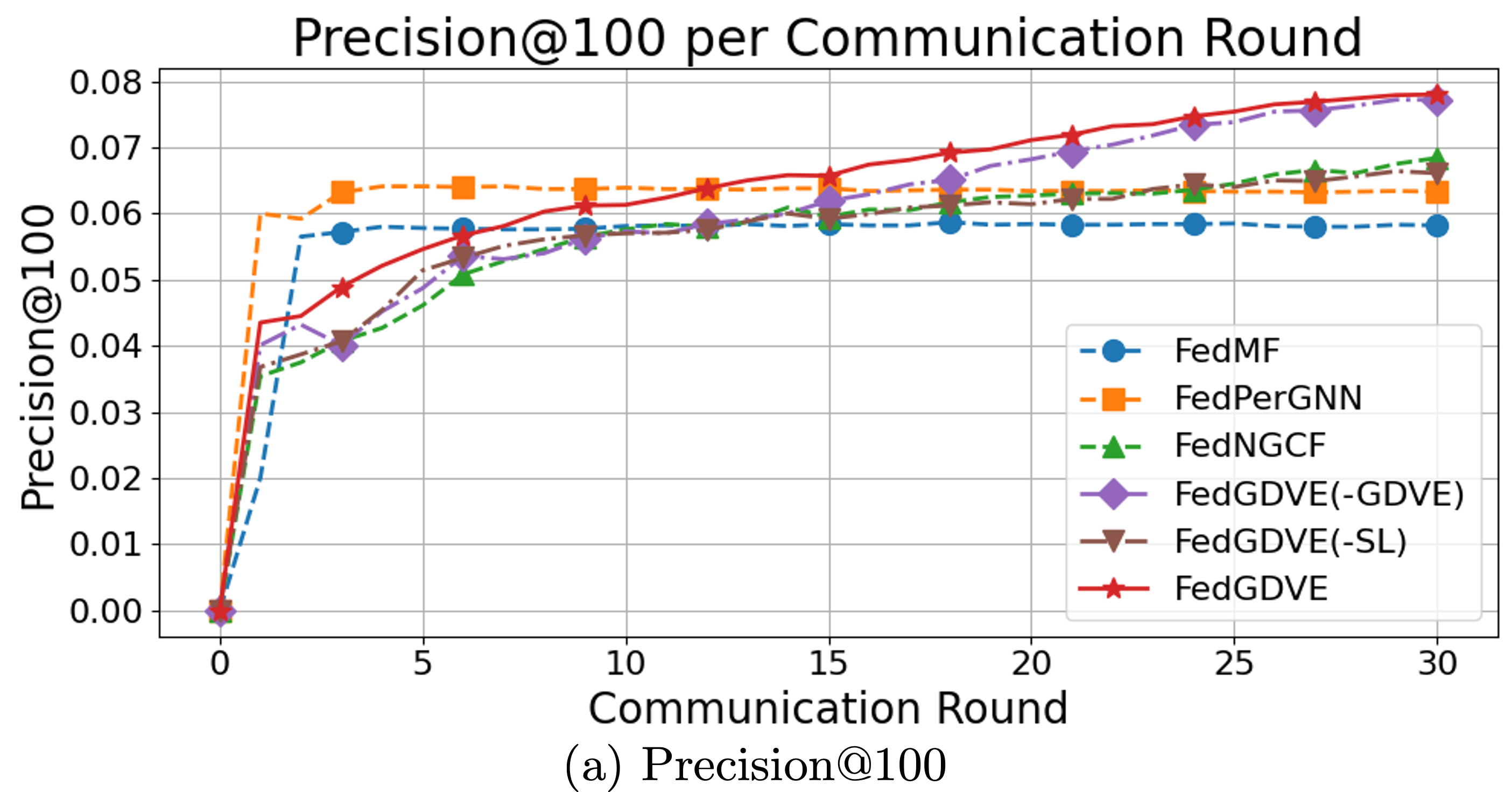}
    \label{fig:precision}
    \includegraphics[width=0.7\linewidth]{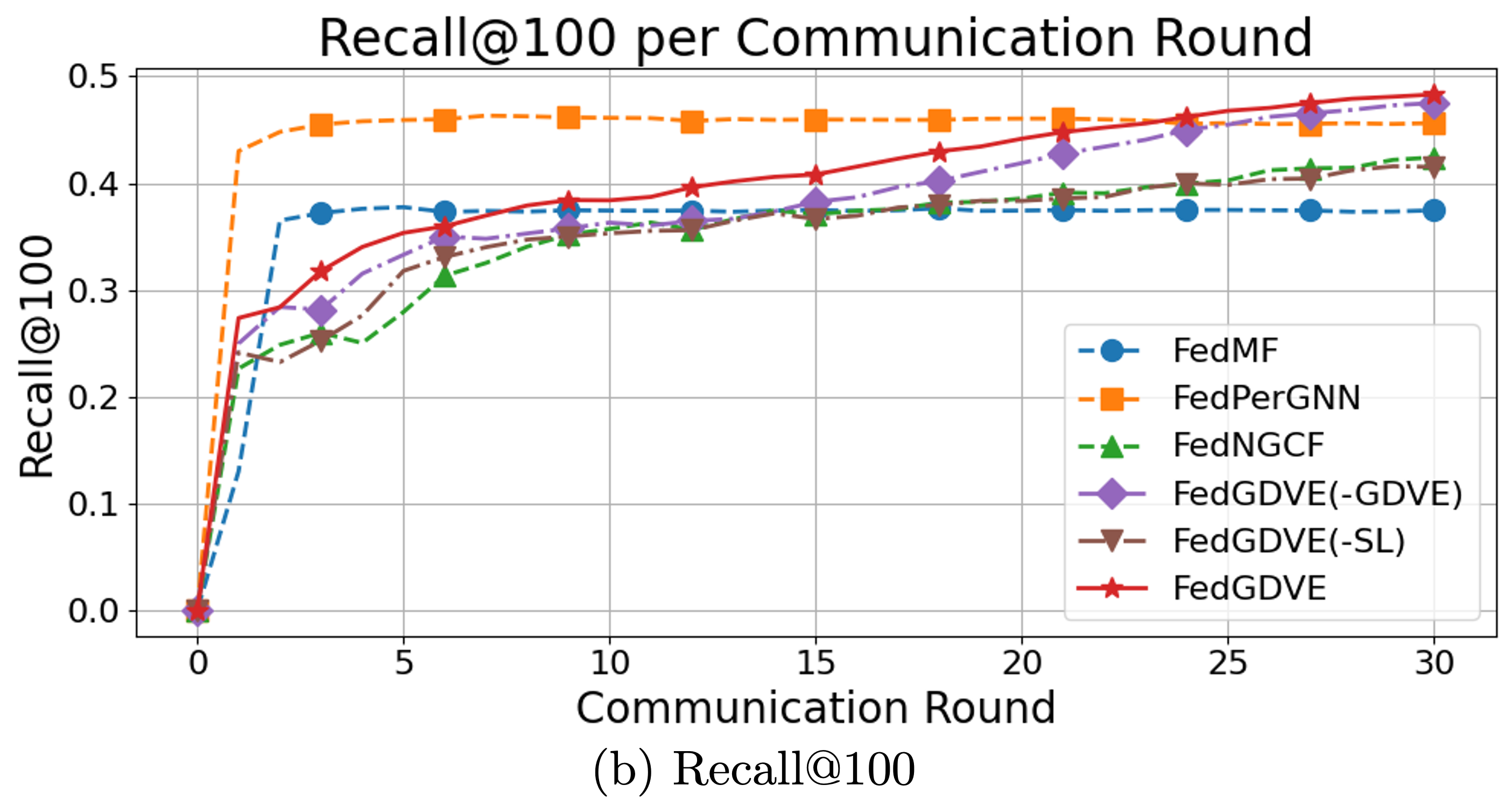}
    \label{fig:recall}
    \includegraphics[width=0.7\linewidth]{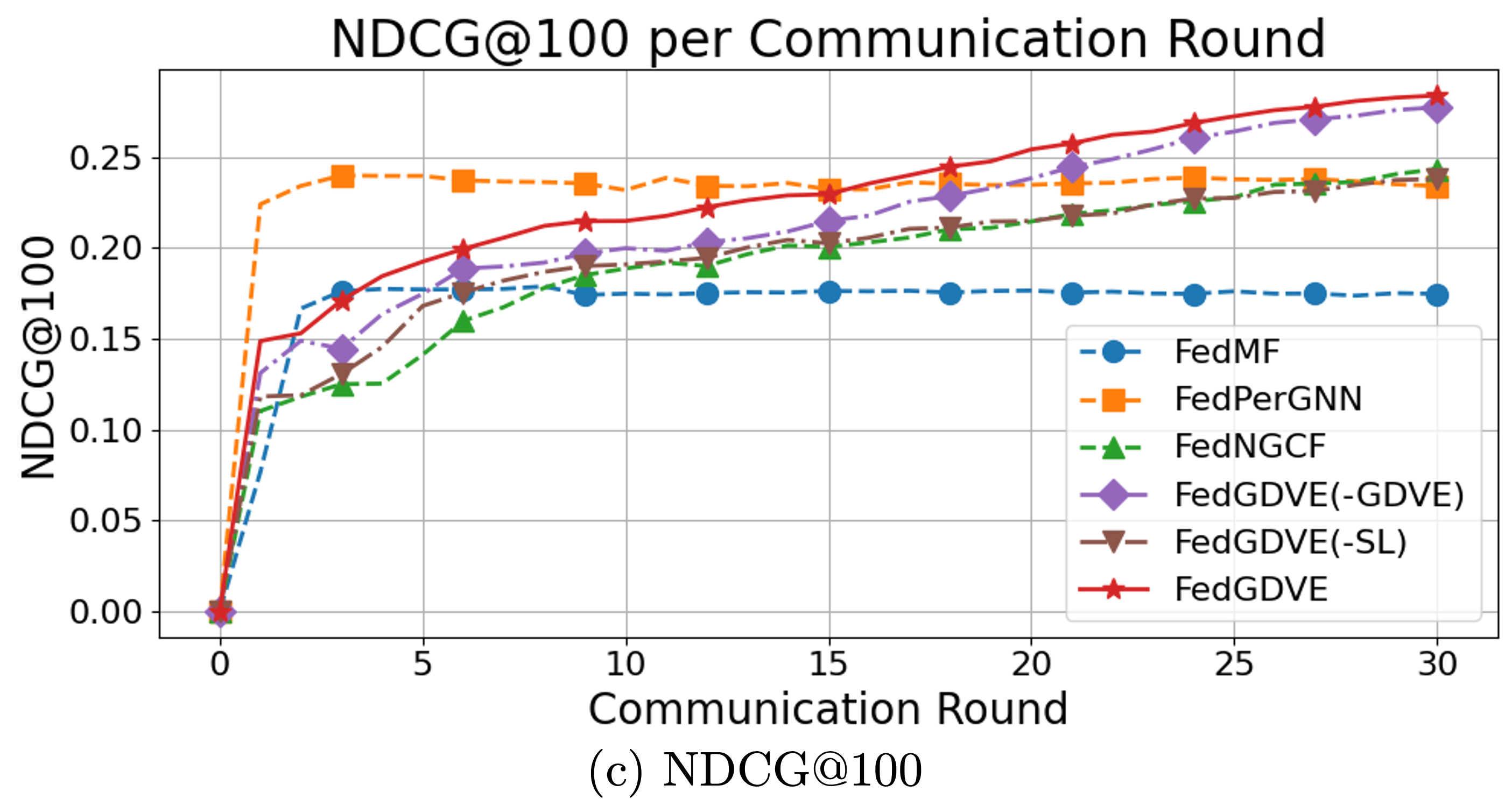}
    \label{fig:ndcg}
    
    \caption{\rev{Comparison of evaluation scores across 30 communication rounds: (a) Precision@100, (b) Recall@100, and (c) NDCG@100. The training of each model proceeds up to the 30th round, based on predefined criteria.}}
    \label{fig:convergence}
\end{figure}

Figure~\ref{fig:convergence} presents a comparative analysis of convergence behavior and recommendation performance between the baseline models and \mname{}, 
\rev{evaluated over 30 communication rounds using the same data partitioning as described in the experimental setup. \mname{} consistently outperforms the baselines, primarily due to its selective data utilization strategy during each training iteration. While \mname{} without the structural loss component (\mname{}(-SL)) exhibits faster convergence, the full \mname{} model achieves superior final performance upon completing the 30 rounds.}



\paragraph{\rev{Influence of Client Number on Performance Variability}}
\begin{table}[!t]
\centering
\caption{Performance based on the number of clients using the MovieLens-100k dataset. The \textbf{boldfaced} score denotes the best result for each number of clients.}
\scriptsize
\begin{tabularx}{\columnwidth}{ll*{4}{Y}}
\toprule
\textbf{Method} & \textbf{Metric} & \textbf{5 Clients} & \textbf{10 Clients} & \textbf{20 Clients} & \textbf{50 Clients} \\
\midrule
\textbf{FedNGCF}   &  & 0.06305 & 0.06302 & 0.06246 & 0.06046 \\
\textbf{FedPerGNN} & \textbf{P@100} & 0.06451 & 0.06407 & 0.06411 & 0.06192 \\
\textbf{FedGDVE}   &  & \textbf{0.08138} & \textbf{0.08499} & \textbf{0.07058} & \textbf{0.07359} \\
\midrule
\textbf{FedNGCF}   &  & 0.46104 & 0.45734 & 0.45494 & 0.44584 \\
\textbf{FedPerGNN} & \textbf{R@100} & 0.46348 & 0.46087 & 0.45928 & 0.45136 \\
\textbf{FedGDVE}   &  & \textbf{0.50263} & \textbf{0.52249} & \textbf{0.46732} & \textbf{0.45981} \\
\midrule
\textbf{FedNGCF}   &  & 0.22735 & 0.22661 & 0.22603 & 0.21486 \\
\textbf{FedPerGNN} & \textbf{N@100} & 0.23330 & 0.23167 & 0.23182 & 0.22236 \\
\textbf{FedGDVE}   &  & \textbf{0.29033} & \textbf{0.31150} & \textbf{0.24841} & \textbf{0.25239} \\
\bottomrule
\end{tabularx}
\label{tab: clients_comparison}
\end{table}

Table~\ref{tab: clients_comparison} presents the performance decrement as the number of clients increases on the MovieLens-100k dataset. This decline can be attributed to the reduced data size for each client and the subsequent increase in data sparsity, which deteriorates the local models' performance. 
\text{FedPerGNN} mitigates this decline to some extent by leveraging 1-hop neighbor information from other clients. However, it is essential to note that the encryption utilized in \text{FedPerGNN} imposes significant computational overhead, particularly as the data volume expands, rendering it impractical for large-scale data scenarios. Finally, \mname{}, which introduces a data selection model, shows the best performance among the three models. By selecting and inserting data tailored to each client, we can achieve more robust training for the shared model. If the number of clients increases, the amount of local data available for learning is reduced, and the performance of \mname{} starts degrading. However, the results of \mname{} are still significantly better than those of the most recent baseline, \text{FedPerGNN}.

\paragraph{\rev{Data Partitioning and Heterogeneity Analysis}}
\begin{table}[!t]
\centering
\caption{Heterogeneity score by partitioning method.}
\small
\resizebox{0.9\linewidth}{!}{%
\begin{tabularx}{\columnwidth}{l*{2}{Y}}
\toprule
\textbf{Distribution} & \textbf{Uniform} & \textbf{Dirichlet} \\
\midrule
Adjusted Mutual Information Score & 0.63 & 0.53 \\
\bottomrule
\end{tabularx}
}
\label{tab:heterogeneity}
\end{table}

A comparative experiment was conducted in both the IID setting, marked by a uniform data distribution among clients, and the non-IID setting, characterized by varying data distributions across clients. To partition the data, we introduce an additional data segmentation technique grounded on node degree. This allows the data to adopt a consistent distribution or display varying distributions. 

Table \ref{tab:heterogeneity} demonstrates the efficacy of the data partitioning based on node degree, evaluated by the adjusted mutual information score. We adopt a uniform distribution for the IID split, whereas a Dirichlet distribution with parameters  \( \alpha=5 \) and \( \beta=9 \) is utilized for the non-IID split. The uniform distribution yields a consistent partition with a score of 0.63. Conversely, partition under the Dirichlet distribution registers a score of 0.53.

\begin{figure*}[!t]
    \centering
    \includegraphics[width=1.0\linewidth]{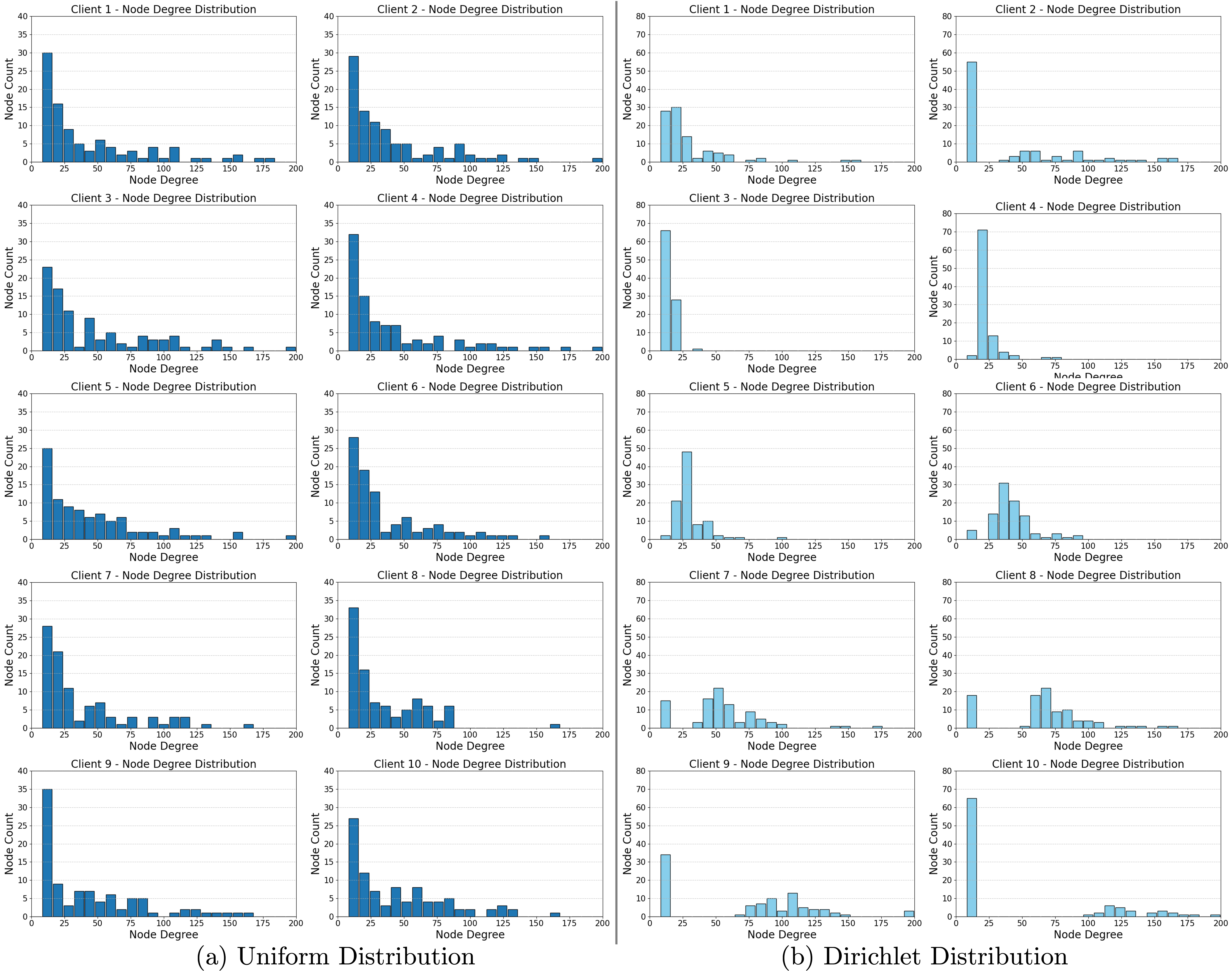}
    \caption{\rev{Comparison of partitioned data distributions. The X-axis is node degree; the Y-axis is node count per degree. (a) Uniform: evenly spread degrees. (b) Dirichlet: skewed degrees with long-tail nodes.}}
    \label{fig:distribution_full}
\end{figure*}

\rev{Figure \ref{fig:distribution_full} illustrates the respective shapes of the data distributions after the split. Figure \ref{fig:distribution_full} (a) reveals uniformity across all nine data partitions, while Figure \ref{fig:distribution_full} (b) depicts considerable variation among them, underscoring the disparity visually. The Dirichlet distribution's parameters afford control over the extent of heterogeneity in the data split.}

\begin{figure*}[!t]
    \centering
    \includegraphics[width=1.0\linewidth]{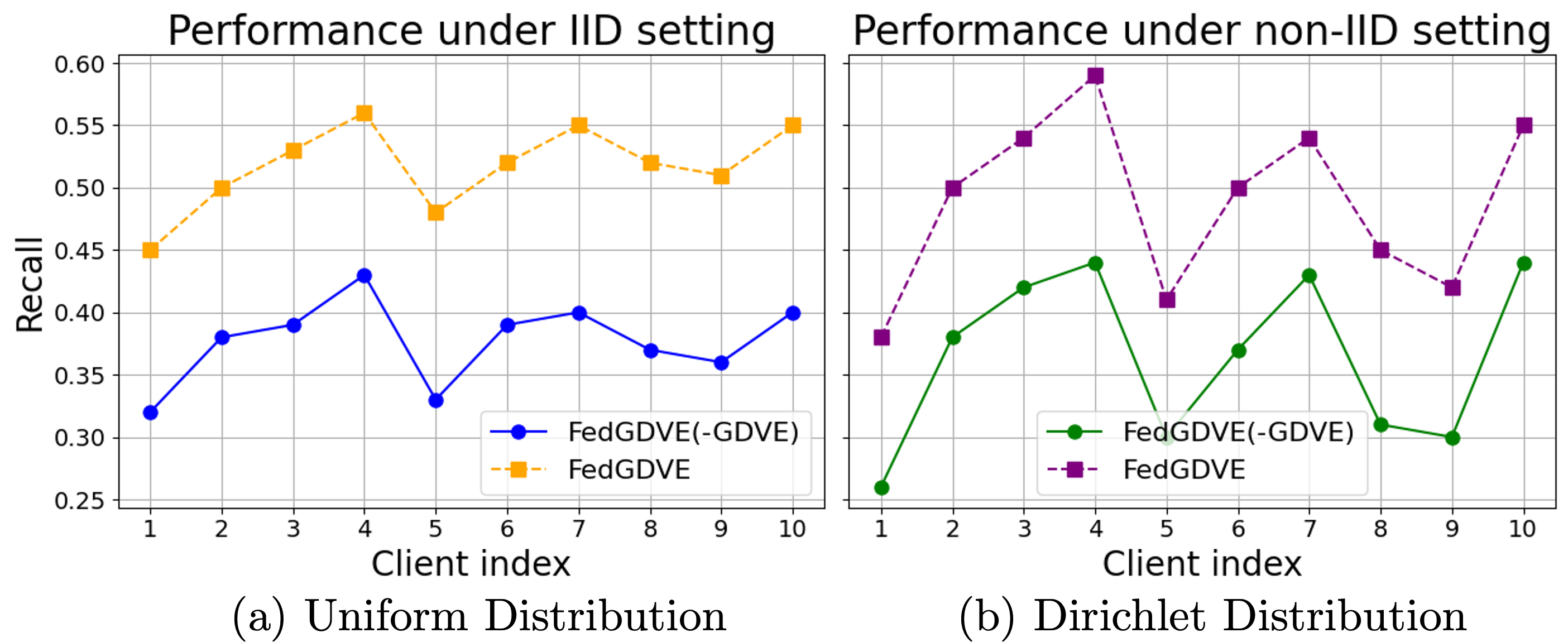}
    \caption{\rev{Performance comparison on two different data distributions.  The X-axis corresponds to an index representing one of the ten clients. The Y-axis represents the recall value measured on the \text{MovieLens-100k} dataset.}}
    \label{fig:data_distribution}
\end{figure*}

\rev{Figure~\ref{fig:data_distribution} presents the outcomes of our federated graph recommender system across ten distinct clients. The Figure~\ref{fig:data_distribution} (a) illustrates performance under an IID data split, while the Figure~\ref{fig:data_distribution} (b) shows results under a non-IID setting with heterogeneous data distributions. Although each client was assigned the same amount of data selected through GDVE, variations in local subgraph structures and user-item interactions naturally lead to performance differences among clients. This variability is expected in federated settings, as each client's model is influenced by its unique data distribution. Notably, our method, \mname{}, consistently outperforms its variant without GDVE across all clients, achieving more than 39\% improvement in recall, with slightly higher gains observed under the non-IID condition.}


\section{Discussion}
\subsection{The challenging cases of varied subgraph structures in global datasets}
\begin{table}[!t]
\caption{Performance evaluation with different global datasets such as MovieLens-100k, Gowalla, and Yelp2018. The MovieLens-100K is chosen as a dataset for local clients. The \textit{Ratio} represents the ratio of interactions within the global dataset selected by GDVE. Items in local clients are virtually connected to the global dataset based on similarity.}
\centering
\resizebox{\textwidth}{!}{%
\begin{tabular}{cccccccccc}
\hline
 & \multicolumn{9}{c}{\textbf{Global dataset}} \\ \cline{2-10} 
 & \multicolumn{3}{c}{\textbf{MovieLens-100k}} & \multicolumn{3}{c}{\textbf{Gowalla}} & \multicolumn{3}{c}{\textbf{Yelp2018}} \\ \hline
\textbf{Local dataset} & \textbf{R@100} & \textbf{N@100} & \textbf{Ratio} & \textbf{R@100} & \textbf{N@100} & \textbf{Ratio} & \textbf{R@100} & \textbf{N@100} & \textbf{Ratio} \\ \hline
\textbf{MovieLens-100k} & 0.2611 & 0.1524 & 11.96\% & 0.1869 & 0.1010 & 2.75\% & 0.2387 & 0.1420 & 2.21\% \\ \hline
\end{tabular}%
}
\label{tab:global_test}
\end{table}

In previous experiments, the global and local datasets were derived from the same source. However, in real-world scenarios, global datasets may display considerable variability in aspects such as item types, connectivity, and node counts.
To assess the performance of \mname{} amidst such diversity, we conducted additional experiments focused on the diversity between local and global datasets. For clients from the MovieLens-100k dataset, global datasets are chosen from the same source, as well as from completely different sources (e.g., Gowalla and Yelp2018). If the global dataset is the same as the client dataset, we arranged the datasets so that there is no overlap in interactions between the local and global datasets. These experiments involve computing the cosine similarity between local users and unknown global items. If the cosine similarity among them surpasses a predefined threshold of 0.4, a virtual link is established. As illustrated in Table \ref{tab:global_test}, when local and global datasets are highly correlated, resembling almost identical datasets, \mname{} selects a significant number of interactions from the global dataset. Conversely, notable differences between the datasets, which reflect a high degree of diversity, result in fewer selected interactions.

Furthermore, although the personal data value estimator effectively selects beneficial global datasets, it does not guarantee consistent state-of-the-art performance when faced with high discrepancies between datasets. This issue could be addressed using strategies such as Data Banzhaf \citep{wang2023data}, which leverages the Banzhaf value from game theory for data valuation. Methods that enhance data security and robustness through Differential Privacy (DP) guarantees and sufficient statistics could also be considered \citep{sim2024incentives}. The Causal Attention Learning (CAL) strategy introduced by \citet{sui2022causal} could also be applied. CAL differentiates causal from non-causal features in GNNs using attention mechanisms and causal theory techniques to enhance generalization and interoperability. Future research can potentially aim to develop a data value estimator that is resilient to the varied characteristics of global datasets, thus addressing the current limitations of \mname{}.

\subsection{Complexity analysis}

We analyze the time complexity of the \mname{} algorithm. Let $\lvert E_{\text{global}} \rvert$ denote the number of edges in the global dataset $D_{\text{global}}$, and $\lvert E_{\text{local}} \rvert$ denote the number of edges in the local dataset. Let $d_1$ refer to the GCN embedding dimension, and $d_2$ refer to the MLP embedding dimension. The time complexity of the centralized GCN is thus expressed as $\mathcal{O}((\lvert E_{\text{global}} \rvert + \lvert E_{\text{local}} \rvert) \cdot d_1^2)$. Considering the pre-trained components, the graph encoder $g_{\varphi}^{k}$ and the validity predictor $f_{\psi}^{k}$, the task predictor $f_{\theta}^{k}$, and the probability estimator $h_{\phi}^{k}$ are pivotal in our complexity analysis. The task predictor's complexity is $\mathcal{O}((S \cdot \lvert E_{\text{global}} \rvert + \lvert E_{\text{local}} \rvert) \cdot d_1^2)$ where $S$ is the selected global data ratio, and the probability estimator, consisting of three MLP layers, has a complexity of $\mathcal{O}(3 \cdot d_2^2)$. Compared to FedNGCF, which has a complexity of $\mathcal{O}((\lvert E_{\text{global}} \rvert + \lvert E_{\text{local}} \rvert) \cdot d_1^2)$, \mname{} exhibits a complexity of $\mathcal{O}((S \cdot \lvert E_{\text{global}} \rvert + \lvert E_{\text{local}} \rvert) \cdot d_1^2 + d_2^2)$.

\begin{table}[!t]
\caption{Using the Yelp2018 dataset, we compare the computational efficiency and training time of \mname{}. The Global Ratio indicates the proportion of the global dataset used, and Time/Epoch refers to the time per epoch. Additionally, the Selected Data Ratio denotes the extent to which each local client selects and utilizes the global dataset.}
\centering
\resizebox{0.8\linewidth}{!}{%
\begin{tabular}{ccccc}
\toprule
\textbf{ } & \textbf{Global Ratio} & \textbf{Time/Epoch} & \multicolumn{1}{l}{\textbf{Selected Data Ratio}} \\ \hline
\multirow{2}{*}{{\textbf{\mname{}}}} & 50\% & 7m 8s & 8.81\% \\ \cline{2-4}
  & 80\% & 10m 30s & 8.92\% \\ \bottomrule
\label{tab:time_complexity}
\end{tabular}
}
\end{table}

Two primary factors influence the time complexity of \mname{}: $S$ and the size of $\lvert E_{\text{global}} \rvert$. These elements are crucial for determining the algorithm's efficiency. Smaller values of $\lvert E_{\text{global}} \rvert$ or the presence of non-IID properties may reduce the sample ratio $S$, thereby enhancing time efficiency. Table \ref{tab:time_complexity} presents a direct comparison of average training time across federated averaging rounds, illustrating the impact of variations in the size of $D_{\text{global}}$. Notably, as the size of the global dataset increases while maintaining a consistent data ratio $S$, the training time tends to increase accordingly.

\clearpage
\subsection{Conclusion}
\rev{This research introduces \mname{}, a new federated learning framework that enhances recommender systems by selectively integrating global data with private client data, addressing the challenge of limited local dataset size and heterogeneity. Utilizing the Graph Data Value Estimator (GDVE), \mname{} filters global user-item interactions to align with each client's data distribution. 
In experiments, \mname{} achieves up to 34.86\% performance improvement over state-of-the-art federated learning methods. While dynamic data selection may slow convergence, the framework demonstrates robust accuracy and scalability across diverse e-commerce scenarios.
Future work will explore subgraph-level federated learning tasks, such as node and link prediction, and investigate server-side client-adaptive data filtering to further optimize learning efficiency and model performance.}





\newpage
\bibliographystyle{model5-names}\biboptions{authoryear}
\bibliography{cas-refs.bib}

\end{document}